%% file: camera-and-arxiv.tex
\documentclass[10pt,twocolumn,letterpaper]{article}

\usepackage[pagenumbers]{cvpr} 
\usepackage{url}

\usepackage{booktabs}       
\usepackage{amsfonts}       
\usepackage{nicefrac}       
\usepackage{microtype}      
\usepackage[dvipsnames]{xcolor}         
\usepackage{amsmath}
\usepackage{multirow}       
\usepackage{graphics}
\usepackage{graphicx}
\usepackage{color, colortbl}
\usepackage{wrapfig,booktabs}
\usepackage{subcaption}
\usepackage{amssymb}
\usepackage{caption}
\usepackage{makecell}
\usepackage{enumitem}
\usepackage{sidecap}        
\usepackage{twemojis}
\usepackage[accsupp]{axessibility}
\input{preamble}

\definecolor{cvprblue}{rgb}{0.21,0.49,0.74}
\usepackage[pagebackref,breaklinks,colorlinks,citecolor=cvprblue]{hyperref}
\def\paperID{8842} 
\def\confName{CVPR}
\def\confYear{2024}

\title{Align before Adapt: Leveraging Entity-to-Region Alignments for
Generalizable Video Action Recognition}

\author{
\textbf{Yifei Chen}
~~~~\textbf{Dapeng Chen}
~~~~\textbf{Ruijin Liu}
~~~~\textbf{Sai Zhou}
~~~~\textbf{Wenyuan Xue}
~~~~\textbf{Wei Peng}\\
IT Innovation and Research Center, Huawei Technologies\\
\{chenyifei14, chendapeng8, liuruijin1\}@huawei.com
}

\begin{document}
\maketitle
\begin{abstract}
Large-scale visual-language pre-trained models have achieved significant success in various video tasks. However, most existing methods follow an ``adapt then align" paradigm, which adapts pre-trained image encoders to model video-level representations and utilizes one-hot or text embedding of the action labels for supervision. This paradigm overlooks the challenge of mapping from static images to complicated activity concepts. In this paper, we propose a novel ``Align before Adapt" (ALT) paradigm. Prior to adapting to video representation learning, we exploit the entity-to-region alignments for each frame. The alignments are fulfilled by matching the region-aware image embeddings to an offline-constructed text corpus. With the aligned entities, we feed their text embeddings to a transformer-based video adapter as the queries, which can help extract the semantics of the most important entities from a video to a vector. This paradigm reuses the visual-language alignment of VLP during adaptation and tries to explain an action by the underlying entities. This helps understand actions by bridging the gap with complex activity semantics, particularly when facing unfamiliar or unseen categories. 
ALT demonstrates competitive performance while maintaining remarkably low computational costs. In fully supervised experiments, it achieves 88.1\% top-1 accuracy on Kinetics-400 with only 4947 GFLOPs. Moreover, ALT outperforms the previous state-of-the-art methods in both zero-shot and few-shot experiments, emphasizing its superior generalizability across various learning scenarios.
\end{abstract}

\begin{figure*}[t]
\centering
\includegraphics[width=0.98\linewidth]{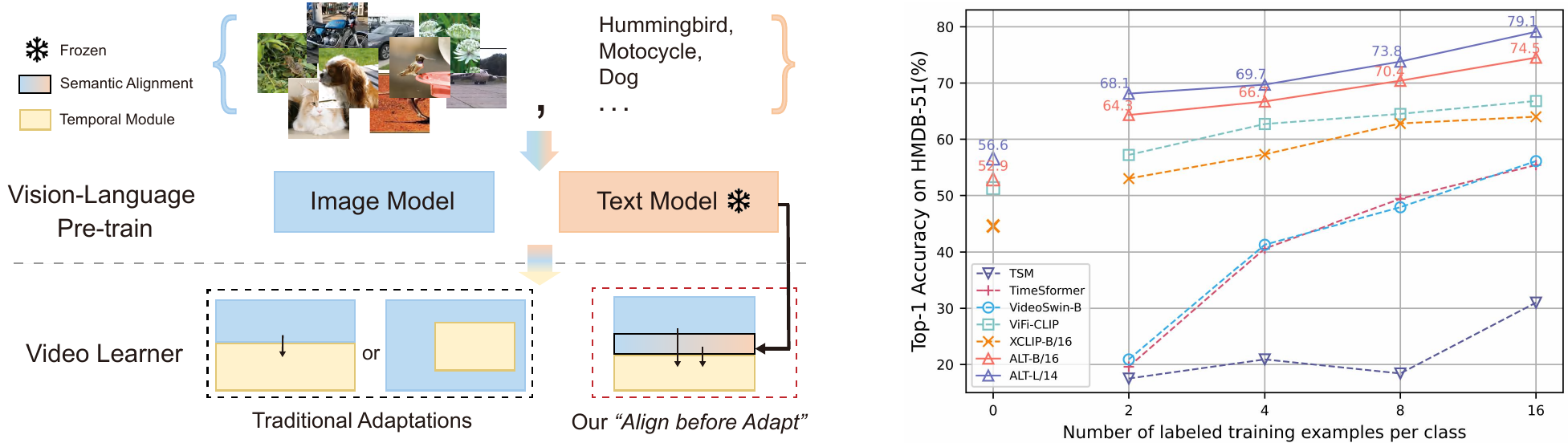}
\caption{\textbf{Left:} Paradigm comparison between traditional adaptation approaches and our ``Align before Adapt'' method.
\textbf{Right:}  Zero-shot and few-shot performance comparison on HMDB-51 dataset. Pretrained on Kinetics-400, our method surpasses the previous state of the arts.} 
\label{fig:align_before_adapt}
 \vspace{-1em}
\end{figure*}

\section{Introduction}
\label{sec:intro}
Video action recognition is a fundamental task in the pursuit of intelligent video understanding. The recent trend of utilizing the visual-language pre-trained (VLP) models~\cite{radfordLearningTransferableVisual2021,chaoALIGN21,YuCoCa22,LiBLIP22} 
have significantly advanced the research of action recognition~\cite{wangActionclipNewParadigm2021,juPromptingVisualLanguageModels2022,panSTAdapterParameterEfficientImagetoVideo2022,niXCLIP,linFrozenCLIPModels2022,YangAIM22}. By lightly fine-tuning the model, VLP-based methods outperform the previous end-to-end network architectures, including two-stream networks~\cite{simonyanTwostreamConvolutionalNetworks2014a, wangTSN, zhouTemporalRelationalReasoning2018}, 3D convolutional neural networks~\cite{carreiraQuoVadisAction2017a,feichtenhoferX3dExpandingArchitectures2020, feichtenhoferSlowfastNetworksVideo2019, haraLearningSpatiotemporalFeatures2017, qiuLearningSpatiotemporalRepresentation2017, tranLearningSpatiotemporalFeatures2015, tranCloserLookSpatiotemporal2018, xieRethinkingSpatiotemporalFeature2018}, 
and vision-transformer-based (ViT) networks~\cite{bertasiusSpacetimeAttentionAll2021, fanMultiscaleVisionTransformers2021, liuVideoSwinTransformer2022a,patrickKeepingYourEye2021, yanMultiviewTransformersVideo2022}. 
Employing a pre-trained VLP model for action recognition can better encode the semantic meaning of items in images, even if they have very different visual appearances. This is very helpful in understanding human action and also explains why VLP models have achieved superior performance. As shown in ~\cref{fig:align_before_adapt}, the current VLP-based action recognition methods follow an ``adapt then align'' paradigm. They either introduce temporal interaction upon image representations or insert temporal modules into pre-trained image encoders. However, the ``adapt then align" paradigm will merely fit the video representation to the action name, which potentially destroys the other visual-semantic correspondences provided by VLP models.  As far as we are concerned, actions are complex concepts that involve multiple fine-grained entities, such as body parts, scenes, and objects. With VLP model, the text embedding of the relevant entities should also be grounded in some image region. ``adapt then align" paradigm does not emphasize the underlying entities-to-regions correspondence behind the action concept. Furthermore, human-centric activities often share common entities, implying that visual-language correspondences can be reused across different actions, even for those that were not included in the training set. The re-usability of entities-to-regions correspondences allow the model to quickly recognize new action categories.

 In this paper, we propose the ``ALign before adapt" (ALT) paradigm. Unlike the ``adapt then align" approaches that align the image-level visual embedding with the text embedding of the action name, ALT aims to establish an entity-level correspondence to support action recognition. The relevant entities should have evidence in specific regions of the image. To achieve entities-to-regions alignment, the VLP model is leveraged in two aspects: (1) Aggregating adjacent and similar image token embeddings from the VLP model. The resulting embedding typically represents a region containing the same entity. (2) Selecting the most relevant entities for each region by matching their image embeddings to the text embeddings of a text corpus.

Using the established alignments, we utilize the text embedding of the entities as queries in a transformer-based decoder for action recognition. This adaptation step helps bridge the gap between general image representations and video action representation while preserving the visual-language correspondences from VLP models. ALT exhibits impressive generalization ability with low computational complexity. It enhances our framework by 6.8\% in top-1 accuracy on the HMDB-51 dataset under the 2-shot configuration, while reducing computational cost by 23\% with the ViT-base backbone. In summary, our contributions are as follows:

\begin{itemize}[leftmargin=*]
\item
We propose an ``align before adapt" paradigm that leverages entities-to-region correspondences to guide the adaptation from VLP to action recognition. The paradigm preserves the visual-language alignment of VLP during the video representation adaption, achieving better interpretability and generalization ability.

\item
We propose a new transformer-based video adapter to extract the semantics of the most important entities from a video to a vector. The adapter adopts a transformer-based architecture, which utilizes the text embedding of the selected entities as the query and the multi-frame visual embeddings as the key and values. 

\item
Extensive experiments under various learning configurations are conducted. Besides demonstrating competitive performance with low computational complexity (surpasses the current leading approach with the same VLP backbone by \emph{0.4\%} top-1 accuracy while requiring \emph{55\%} fewer GFLOPs). Our method reveals superior generalizability due to the reusable text entities
(surpasses the previous state-of-the-art by more than \emph{10}\%).
\end{itemize}

\begin{figure*}[t]
\centering
\includegraphics[width=0.95\linewidth]{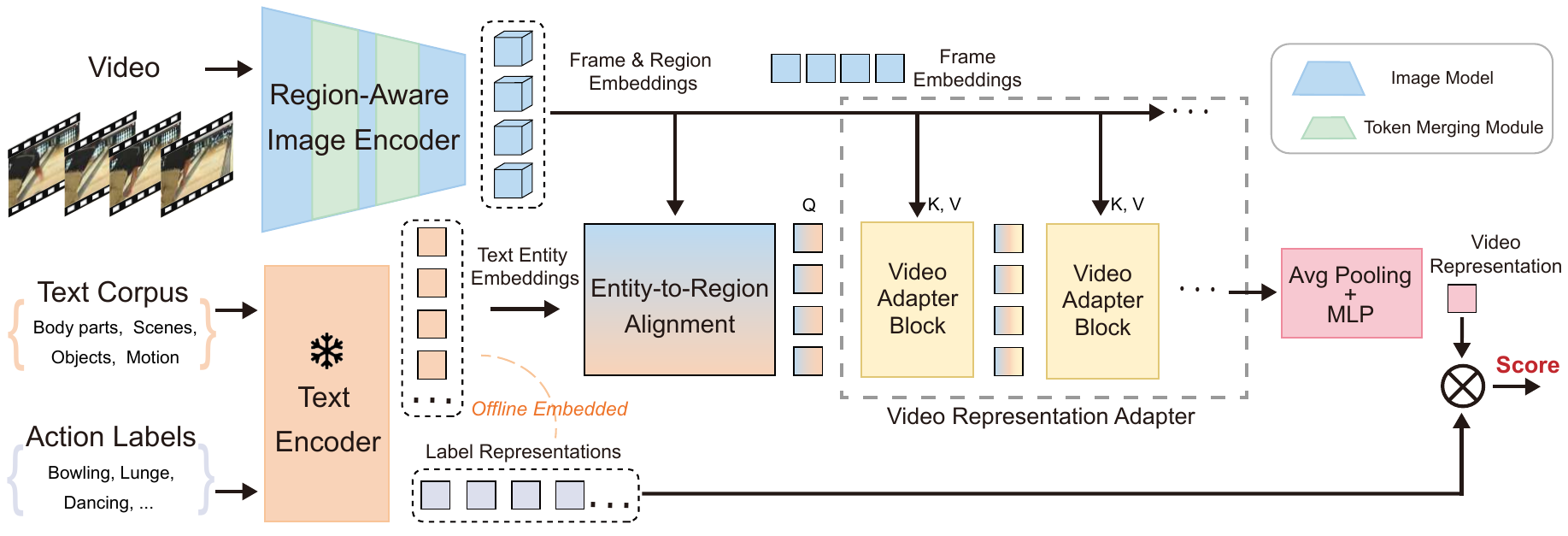}
\caption{An overview of our framework: 
we utilize a video clip and an offline text corpus as inputs to learn a \textcolor{CarnationPink}{video representation}, which is supervised with the objective of maximizing the similarity score with the text representation of the corresponding action label.
} 
\label{fig:network} \vspace{-1em}
\end{figure*}
\section{Related Work}

\noindent \textbf{Large Scale Visual-Language pretraining.}
In the past few years, the surge of large-scale visual-language pre-trained (VLP) models ~\cite{SuVLBert20,liOscarObjectsemanticsAligned2020,radfordLearningTransferableVisual2021,chaoALIGN21,LiBLIP22, ZengAligningTextswithVisual22} 
have revolutionized multiple fields of computer vision, including image classification,
captioning~\cite{YuCoCa22},
grounding~\cite{LiGLIP22},
image-text retrieval, and so on.
With the availability of massive amounts of web-scale visual-text paired data, these models learn cross-modal representations through masked language modeling and/or contrastive learning.
Specifically, one of the most representative works, CLIP ~\cite{radfordLearningTransferableVisual2021}
, is trained on 400M data following a contrastive manner, and shows remarkable performance on zero-shot image classification. 
The success of VLP models inspires the ``fine-tuning'' trend on multiple downstream tasks, such as open-vocabulary detection~\cite{guOpenvocabularyObjectDetection2021}, segmentation~\cite{wangCrisClipdrivenReferring2022,XuODISE23,GhiasiOpenSeg22}, caption~\cite{mokadyClipcapClipPrefix2021a}, summarization~\cite{narasimhan_clip-it_2021}, generation~\cite{AdityaDALLE2_22}, etc. 
Our method adopts CLIP as the backbone for video action recognition tasks under fully-supervised, few-shot, and zero-shot scenarios.

\noindent \textbf{Video Action Recognition.}
The prosperity of deep learning has sparked various works for effective video action recognition. Initially, there were two directions of methods: two-stream 2D CNNs~\cite{zhouTemporalRelationalReasoning2018, wangTSN, simonyanTwostreamConvolutionalNetworks2014a} 
that process and spatial and temporal context parallelly, and 3D CNNs~\cite{
tranLearningSpatiotemporalFeatures2015, 
tranCloserLookSpatiotemporal2018, qiuLearningSpatiotemporalRepresentation2017,
xieRethinkingSpatiotemporalFeature2018, feichtenhoferSlowfastNetworksVideo2019, feichtenhoferX3dExpandingArchitectures2020} that factorize the convolution across spatial and temporal dimensions simultaneously.
Later transformer-based approaches, including 
ViViT~\cite{arnabVivitVideoVision2021a}, Timesformer~\cite{bertasiusSpacetimeAttentionAll2021}, 
and VideoSwin~\cite{liuVideoSwinTransformer2022a}
outperformed the convolutional methods,
by better capturing long-term dependencies through scalable self-attention mechanisms.
More Recently, leveraging available VLP models such as CLIP~\cite{radfordLearningTransferableVisual2021}
and Florence~\cite{LuFlorence}, 
has become a data-friendly trend. 
EVL~\cite{linFrozenCLIPModels2022}, 
ST-Adapter~\cite{panSTAdapterParameterEfficientImagetoVideo2022}, 
and AIM~\cite{YangAIM22} 
add lightweight modules to the fixed CLIP backbone for close-set recognition tasks, while ActionCLIP~\cite{wangActionclipNewParadigm2021}
, X-CLIP~\cite{niXCLIP},
and ViFi-CLIP~\cite{RasheedViFiCLIP23}
propose frameworks that enable adaptation to new scenarios.
While the above methods focus on adapting the visual branch of models to video input directly, our approach explores `entity-to-region' visual-semantic alignments before the adaptation. This bridges the gap of mapping with complicated activity semantics during video representation learning.

\noindent \textbf{Region-Aware Perception for Vision Transformer.}
In recent research on ViT architectures, it has been well-studied that capturing fine-grained patterns in visual signals leads to improved representation learning. Various  approaches, such as 
Swin Transformer~\cite{liuVideoSwinTransformer2022a}, 
Region ViT~\cite{ChenRegionViT22}, 
and GCViT~\cite{AliGCViT23}, 
propose incorporating multi-scale attention into the ViT to achieve better performance in various downstream tasks including recognition, detection, and segmentation. 
With the rise of visual-language pretraining, GLIP~\cite{LiGLIP22} suggests learning better instance-level language-aware representations through grounded image-text data, while FILIP~\cite{YaoFILIP22} and Dense CLIP~\cite{RaoDenseCLIP22} focus on introducing patch-level contrastive losses. These works have shown impressive progress in various scenarios. 
In contrast to methods that require additional structures, data, or supervision, we adopt ToMe~\cite{BolyaToMe22}, which initially aims at boosting ViTs with minor performance drops, in our image encoder. Finding its soft bipartite matching strategy tends to cluster patch tokens of frames with similar region/instance representativeness, we utilize the merged tokens to achieve ``entity-to-region'' visual-semantic alignments for video action recognition.  


\begin{figure*}[t]
\centering
\includegraphics[width=0.95\linewidth]{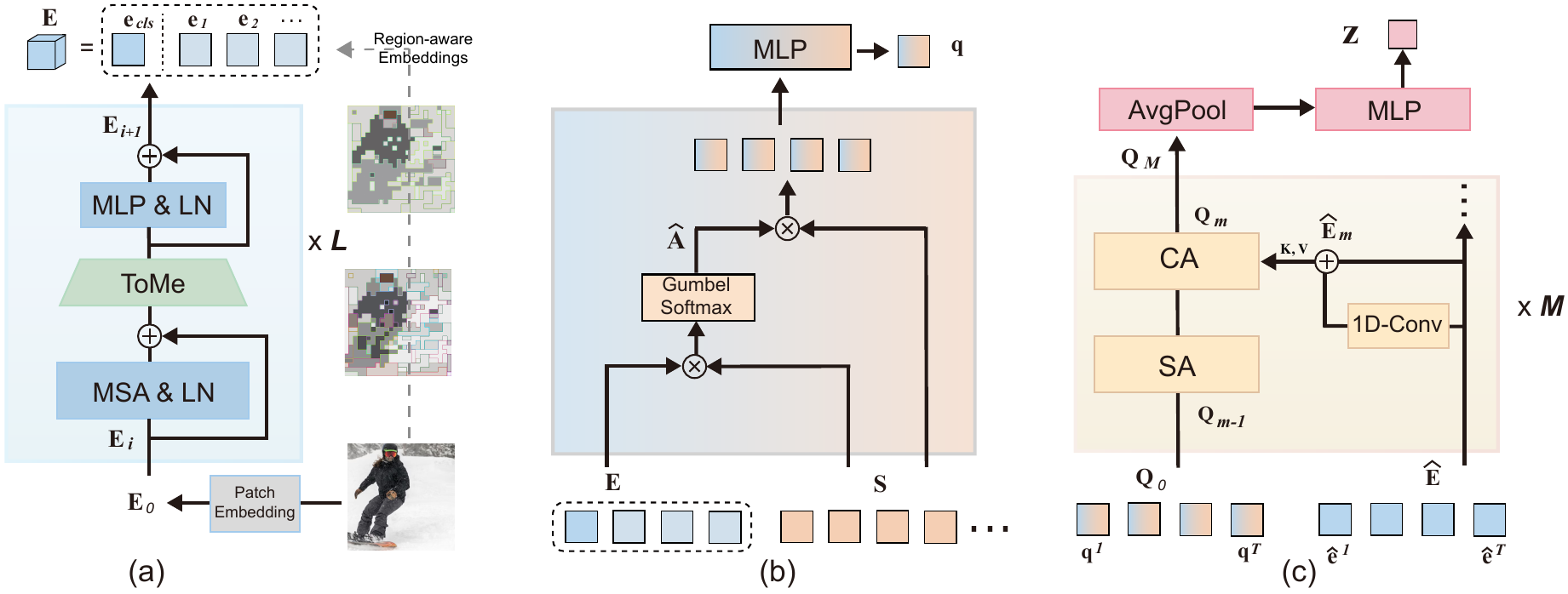}
 \vspace{-0.6em}
\caption{Detailed network components:
(a) The region-aware image encoder includes a ViT with plug-in token merging modules. \textit{MSA}, \textit{MLP}, and \textit{LN} indicate multi-head self-attention, multilayer perception, and layernorm, respectively. 
(b) The entity-to-region alignment module obtains the aligned query in a softmax-weight-sum manner.
and (c) shows the multi-modal video adapter, with each block containing a stack of hybrid modules composed of attention layers and 1D temporal convolution.} 
\label{fig:network_detail} \vspace{-1em}
\end{figure*}

\section{Methodology}


Our method proposes an ``align before adapt'' framework to learn discriminative and transferable video representation. The overview of our framework is illustrated in \cref{fig:network}. The framework has two main steps. First, we explore the entities-to-regions alignment from a constructed text-corpus and the region-aware image embeddings (\cref{sec:Visual_semantic_alignment}). Then we leverage the text embeddings of aligned entities to guide the adaption to the video representation (\cref{sec:video_decode}).


\subsection{Entitiy-to-Region Alignments} \label{sec:Visual_semantic_alignment}

\noindent \textbf{Text corpus of action-related entities.} Drawing inspiration from cognitive science and recent research~\cite{kurbySegmentationPerceptionMemory2008,zacksPerceivingRememberingCommunicating2001, fangPairwiseBodyPartAttention2018, liPastanetHumanActivity2020}, 
we believe that perceiving and leveraging intermediate spatiotemporal-variated patterns, such as bodies, objects, and scenes, can greatly mitigate the difficulty of understanding activity concepts. Thanks to the VLP models, these patterns can be linguistically expressed and perceived based on their similarities with visual representations in the embedding space. We construct a knowledge base for these patterns, referred to as ``text corpus''.

To generate a reusable text corpus, we first collect a set of action labels from several recognition datasets including ~\cite{joaok600, HildegardHMDB51, KhurramUCF101}. The action-related text corpus is then generated with our proposed automatic pipeline: 
(1) Extracting and expanding entities from original action names
by utilizing POS (Nouns) tagging tools~\cite{NLTK,spacy2} and prompting LLM~\cite{OpenAIChatGPT22}. 
(2) Generating a set of descriptions for each entity by utilizing WordNet~\cite{GeorgeWordNet} and LLM. 
(3) Employing word sense disambiguation techniques, including the Lesk algorithm~\cite{basile2014EnhancedLesk} and T5 language model~\cite{wahle2021WSDinLM}, to filter out unrelated descriptions of an entity to action. In addition, body parts, such as \texttt{head} and \texttt{feet} are added to the corpus by default. They are involved in most human activities. The details and utilized prompts are provided in supplementary materials. 

The text corpus, denoted as $\mathcal{S} = \{ s_{i}\}_{i=1}^{K}$, consists of K entities, where each entity $s_{i}$ is represented by its entity name and description: $s_{i}$ = \{\texttt{unit}$_{i}$, \texttt{description}$_{i}$\}. The text embeddings for all the text entities can be denoted as $\textbf{S}\in \mathbb{R}^{K \times d}$, where d represents the dimension of embeddings.


\noindent \textbf{Region-aware image embeddings.} To fully explore the entity-to-region alignments, we need to perceive the region-level image embeddings. 
With a CLIP-ViT, the frame input is initially divided into $N$ non-overlapping patches and projected to a sequence of $d$-dimension embeddings $\mathbf{E}_{0}$:
\begin{equation}
 \textbf{E}_{0} = [\textbf{e}_{cls,0}, \textbf{e}_{1,0}, \textbf{e}_{2,0}, ..., \textbf{e}_{N,0}] + \textbf{E}_{pos},
\end{equation}
where each embedding $\mathbf{e}_{n,0}$ corresponds to a patch. $\mathbf{e}_{cls,0}$ and $\mathbf{E}_{pos}$ denote prepended $[class]$ embedding and position embeddings, respectively.
However, the image patches are too redundant to represent a region. We adopt token-merging~\cite{BolyaToMe22} to aggregate patches. The token merging technique originally aimed to accelerate the ViT architecture.  It employs a soft bipartite matching algorithm to find the most similar $r$ pairs of embeddings (excluding $[cls]$). Each pair of embeddings are merged into one, reducing the total number by $r$.  The token merging module is inserted into each transformer block, denoted by \texttt{BLOCK}$_{ToMe}$. The forward pass of the modified transformer block can be formulated as:
\begin{equation}
  \mathbf{E}_{l} = \texttt{BLOCK}_{ToMe} (\mathbf{E}_{l\!-\!1}),
\end{equation}
where $l$ denotes the index of transformer blocks, and $\mathbf{E}_{l}\in \mathbb{R}^{(N+1-l\times r)\times d}$ reduces $r$ patch embeddings compared with $\mathbf{E}_{l-1}$. After passing through $L$ blocks (the second last layer of ViT), the final image embeddings are represented by:
\begin{equation}
    \mathbf{E} = [\mathbf{e}_{cls}, \mathbf{e}_{1}, \mathbf{e}_{2},...,\mathbf{e}_{N-L\times r}],  \label{eq:image_embedding}
\end{equation}
where $\mathbf{e}_{1}, \mathbf{e}_{2},...,\mathbf{e}_{N-L\times r}$ are region-aware embeddings that account for corresponding merged patches which have similar visual representativeness. We visualize the merging procedure along the transformer layers in Fig.~\ref{fig:network_detail}a, where the patches with the same color and border are merged into one and form region-aware embeddings.

\noindent \textbf{Entity-to-regions alignments.}  ALT aims to build the entities-to-regions correspondences by matching the text embeddings of entities and the region embeddings of each frame. As depicted in \cref{fig:network_detail}b, the process starts by calculating the similarity matrix $\mathbf{A}$ between the visual embeddings $\mathbf{E}$ and text entity embeddings $\mathbf{S}$ through Gumbel-Softmax~\cite{JangCategoricalReparameterizationGumbel-Softmax17,MaddisonContinuousDiscreteRandomVariables17} over $\mathbf{S}$ 
with $Gumbel(0, 1)$ distributed samples ${\gamma}$, where:
\begin{equation} \label{eq:semantic_attention}
\mathbf{A}_{i,j}= \frac
{\exp(\langle \mathbf{e}_i, \mathbf{s}_j\rangle
/
(\|\mathbf{e}_i\|\cdot\|\mathbf{s}_{j}\|)+\gamma_j)}
{\sum_{k=1}^K 
\exp(\langle \mathbf{e}_i, \mathbf{s}_k\rangle
/
(\|\mathbf{e}_i\|\cdot\|\mathbf{s}_{k}\|)+\gamma_k)}.
\end{equation}
Note that $\mathbf{e}_{i} \in \mathbb{R}^{d}$ and $\mathbf{s}_{j} \in \mathbb{R}^{d}$ are $i$th and $j$th embeddings of $\mathbf{E}$ and $\mathbf{S}$, respectively.

To address the ambiguities of aligned semantics, we 
highlight the most similar entity for each region 
by enforcing a differentiable one-hot trick~\cite{OordVQVAE17,XuGroupViT22}:  
\begin{equation} \label{eq:one_hot_operation}
\widehat{\mathbf{A}}=one\mbox{-}hot(\mathbf{A}_{argmax})+\mathbf{A}-detach(\mathbf{A}),
\end{equation}
where $detach$ stops the gradient. 
$\widehat{\mathbf{A}} \in \mathbb{R}^{(N+1-L\times r)\times K}$ only aligns the most correlated text entity for each frame-level or region-aware embedding with dominating weights while keeping differentiable.
It is noteworthy that, to validate the precision and interpretability of visual-semantic alignments, in \cref{fig:align_vis_and_GFLOPs_compare} left, we visualize the correspondence between region-aware embeddings and text entities. 

\subsection{Video Representation Adapter}\label{sec:video_decode}

The video adapter aims to extract the most discriminative information from multi-frame visual embeddings to a single vector for action recognition. Previous methods utilize the labels or the text embeddings of the action names to supervise the adapter, while our method can leverage the relevant entities obtained from the entity-to-region alignments. 

More specifically, we utilize the transformer-based architecture for the adapter. 
The queries, keys, and values are reduced embeddings of matched entities and video frames. 
Take one frame as an example: query $\mathbf{q}$ is calculated by first summing the text embedding $\mathbf{S}$ weighted by the similarity matrix $\widehat{\mathbf{A}}$ then reducing to a vector by MLP. Meanwhile, key and value are the same vector $\widehat{\mathbf{e}}$, obtained by processing the image embedding $\mathbf{E}$ with another MLP:
\begin{equation}
    \mathbf{q} =  {\rm MLP}({\widehat{\mathbf{A}}\mathbf{S}}), 
    \quad
    \widehat{\mathbf{e}} =  {\rm MLP}(\mathbf{E}).
    \label{eq:query} 
\end{equation}
Given a video with $T$ frames $[\mathbf{I}^{1}, \mathbf{I}^{2},...,\mathbf{I}^{T}]$, the initial
queries $\mathbf{Q}_{0}$ to the adapter are query vectors of different frames, while the keys and values are $\widehat{\mathbf{E}}$, 
which is the processed image embeddings of input frames (Eq. \ref{eq:query}):
\begin{equation}
    \begin{split}
        \mathbf{Q}_{0} \!=\! 
        [\mathbf{q}^{1},...,\mathbf{q}^{T} ],
        \quad
        \widehat{\mathbf{E}} \!=\! 
        [\widehat{\mathbf{e}}^{1},...,\widehat{\mathbf{e}}^{T}], 
    \end{split}
\end{equation}
The structure of our adapter is illustrated in \cref{fig:network_detail}c. This adapter includes a sequence decoding block that consists of a 1D-convolution module, a self-attention (SA) module, and a cross-attention (CA) module. The attention modules function in the same way as the ones in the transformer~\cite{AttentionAllYouNeed}. The procedure of the video adapter can be formulated as:
\begin{equation}
\begin{split}
     & \mathbf{Q}_{m}^{\prime} = {\rm SA}_{m}(\mathbf{Q}_{m-1}), \quad
      \widehat{\mathbf{E}}_{m}  =  \widehat{\mathbf{E}} + {\rm 1D\mbox{-}Conv}_{m}(\widehat{\mathbf{E}}),
     \\
     & \mathbf{Q}_{m} = {\rm CA}_{m}(\mathbf{Q}_{m}^{\prime}, \widehat{\mathbf{E}}_{m}),\quad m = 1,...,M,
\end{split}
\end{equation}
where $m$ indicates the block index of the video adapter. 
The SA module and the 1D-convolution module serve for temporal interactions for $\mathbf{Q}_{m-1}$ and $\widehat{\mathbf{E}}$, respectively. The CA module utilizes evolved text embeddings as queries to aggregate to the visual embeddings across the frames, preserving the entity-level visual information during the adaption. The output query after $M$ blocks is $\mathbf{Q}_{M}$.  
We obtain the final video representation $\mathbf{z}$ by applying Average Pooling and MLP layer over the $\mathbf{Q}_{M}$ sequentially along the temporal dimension and the feature channel:
\begin{equation}
\label{eq:video_represent_out}
    \mathbf{z} = {\rm MLP}({\rm AvgPool}(\mathbf{Q}_{M})).
\end{equation}

\subsection{Training Details} \label{sec:training_details}
\noindent \textbf{Loss function.}
Our proposed network aims to maximize the similarity between video representations and textual representations of corresponding action labels. Specifically, we utilize the frozen text encoder of CLIP to perform prompt ensembling for action labels with a bunch of handcraft templates~\cite{niXCLIP}. Given the representations of the $i$th action $\mathbf{c}_{i}$ and $n$th video $\mathbf{z}_{n}$ (as described in \cref{eq:video_represent_out}), the loss function can be implemented by the cross-entropy loss:
\begin{equation}
    \mathcal{L} = -\frac{1}{N}\sum_{n=1}^{N}\sum_{i=1}^{I} y^{i,n} \log \left( \frac{\exp(\mathbf{c}_{i}^{\top}\mathbf{z}_{n})}{\sum_{j=1}^{I} \exp(\mathbf{c}_{j}^{\top}\mathbf{z}_{n})} \right).
\end{equation}
The training set has $N$ videos belonging to the $I$ actions. 
If the $n$th video belongs to the $i$th action, $y^{i,n}$ equals 1; otherwise, $y^{i,n}$ equals 0.

\noindent \textbf{Network training.}
The ViT backbone in the region-aware image encoder is initialized by CLIP, while the token merging module is parameter-free with the reduction number $r$ to be 8.
The number of blocks in the multi-modal video decoder is set to 4 and 6 for ViT-B and ViT-L backbones, respectively. 
We adopt an AdamW optimizer for network parameter training with initial learning rates of 8$\times10^{-6}$ for the ViT backbone and 8$\times10^{-5}$ for the remaining parts.
The networks are trained with 30 epochs (including a five-epoch warmup) and a weight decay of 0.001 w.r.t. a cosine schedule.
The input video follows the main sparse sampling method~\cite{wangTSN} and augmentation strategy~\cite{niXCLIP} with a frame resolution 224$\times$224. 
Experiments are conducted with 8 32GB-V100-GPUs. 

\definecolor{brilliantlavender}{rgb}{0.96, 0.73, 1.0}
\definecolor{celadon}{rgb}{0.67, 0.88, 0.69}
\definecolor{columbiablue}{rgb}{0.61, 0.87, 1.0}
\definecolor{azure(web)(azuremist)}{rgb}{0.94, 1.0, 1.0}
\definecolor{lavender(web)}{rgb}{0.9, 0.9, 0.98}
\definecolor{lavenderblue}{rgb}{0.8, 0.8, 1.0}
\definecolor{lavendermist}{rgb}{0.92, 0.9, 0.98}
\definecolor{mistyrose}{rgb}{1.0, 0.89, 0.88}
\definecolor{bubbles}{rgb}{0.89, 0.98, 1.0}
\definecolor{bubblegum}{rgb}{0.99, 0.76, 0.8}
\definecolor{babyblueeyes}{rgb}{0.63, 0.79, 0.95}
\begin{table*}
\centering
\setlength{\tabcolsep}{3.3mm}
\resizebox{0.9\linewidth}{!}{
\begin{tabular}{lccccccc}
\toprule
Method & 
Pretrain & Top-1 & Top-5 & GFLOPs & Frames & Views &  $\#$Param.(M)  \\
\midrule
\multicolumn{7}{l}{\textit{Methods with ImageNet or web-scale image pretraining}} \\
MViTv1-B~\cite{fanMultiscaleVisionTransformers2021} 
& -   & 81.2  & 95.1 & 4095 & 64 & 3$\times$3 & 36.6  \\
Uniformer-B~\cite{kunchangUniFormer}
& IN-1k   & 83.0  & 95.4 & 3108 & 32 & 4$\times$3 & 50.0  \\
TimeSformer-L~\cite{bertasiusSpacetimeAttentionAll2021}
& IN-21k  & 80.7  & 94.7 & 7140 & 64 & 1$\times$3 & 121.4 \\
VideoSwin-L~\cite{liuVideoSwinTransformer2022a}
& IN-21k  & 83.1  & 95.9 & 7248 & 32 & 4$\times$3 & 200.0  \\
ViViT-H/16~\cite{arnabVivitVideoVision2021a}
& JFT-300M  & 84.9  & 95.8 & 48916 & 32 & 4$\times$3 & 647.5 \\
\midrule
\multicolumn{7}{l}{\textit{Methods with web-scale language-image pretraining}}  \\
\rowcolor{bubbles}
MTV-H/16~\cite{yanMultiviewTransformersVideo2022}
& WTS-17B* & 89.1  & 98.2 & 45537 & 32 & 4$\times$3  & -  \\
PromptingCLIP-B/16~\cite{juPromptingVisualLanguageModels2022}
& CLIP-400M  & 76.9  & 93.5  & - & 16 & 5$\times$1 & 95.5    \\
ActionCLIP-B/16~\cite{wangActionclipNewParadigm2021}
& CLIP-400M  & 83.8  & 97.1  & 16890 & 32 & 10$\times$3 & 105.2  \\
ViFi-CLIP~\cite{RasheedViFiCLIP23}
& CLIP-400M   & 83.9  & 96.3 & 3372  & 16 & 4$\times$3  & 124.7 \\
ASU-B/16~\cite{Chen22ASU}
& CLIP-400M   & 84.7  & 96.8 & 3444  & 16 & 4$\times$3  & 132 \\
ST-Adapter-L/14~\cite{panSTAdapterParameterEfficientImagetoVideo2022} 
& CLIP-400M   & 87.2  & 97.6 & 8248  & 32 & 3$\times$1  & - \\
EVL-L/14~\cite{linFrozenCLIPModels2022}
& CLIP-400M   & 87.3  & 97.6 & 8088  & 32 & 3$\times$1  & 357.9 \\
AIM-L/14~\cite{YangAIM22}
& CLIP-400M  & 87.5  & 97.7 & 11208 & 32 & 3$\times$1  & 341\\
X-CLIP-L/14~\cite{niXCLIP}
& CLIP-400M  & 87.1  & 97.6  & 7896 & 8 & 4$\times$3  & 451.2\\
X-CLIP-L/14(336$\uparrow$)$\dagger$ ~\cite{niXCLIP}
& CLIP-400M  & 87.7  & 97.4  & 37032 & 16 & 4$\times$3  & 451.2\\
\midrule
\multicolumn{7}{l}{\textit{Our method}}  \\
ALT-B/16
& CLIP-400M    & 84.8  & 96.4  & 657 & 16 & 3$\times$1  & 134.4 \\
ALT-B/16
& CLIP-400M  & 85.5  & 96.7  & 1308 & 32 & 3$\times$1  & 134.4 \\
ALT-L/14
& CLIP-400M   & 87.8  & 97.6  & 2478 & 16 & 3$\times$1  & 437.1\\
\rowcolor{mistyrose}
ALT-L/14
& CLIP-400M  & 88.1  & 97.7  & 4947 & 32 & 3$\times$1  & 437.1\\
\bottomrule
\end{tabular}
}
\vspace{-1em}
\caption{Comparison to state-of-the-art on Kinetics-400. Views are denoted with \textit{``temporal clips $\times$ spatial crops.''}
Parameters in text encoders are not counted.
\** indicates pretraining with a video-text collection.
$\dagger$ indicates the input frame size is 336$\times$336.}
\label{tab:k400}
\end{table*}

\begin{figure*}[t]
\hfill
\begin{minipage}[t]{0.63\textwidth}
    \centering
    \vspace{-0.5em}
    \includegraphics[width=0.95\linewidth]{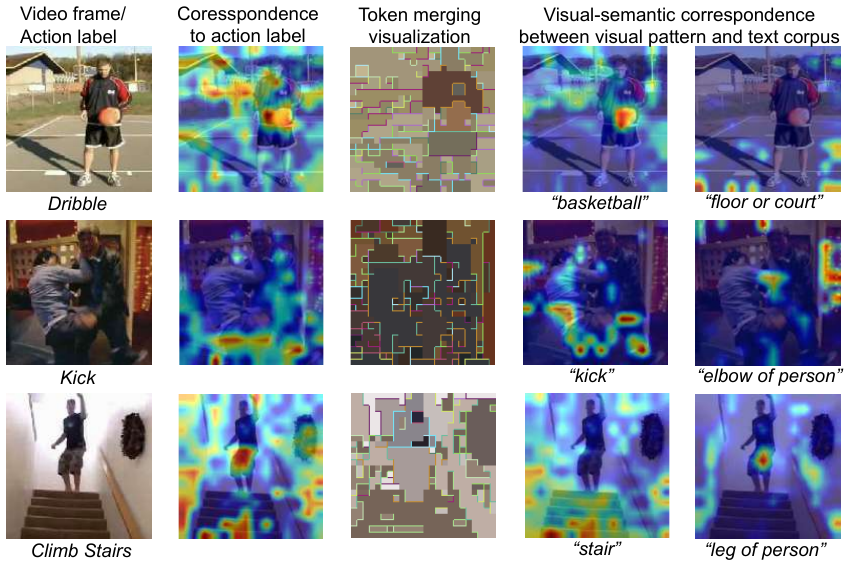}
    \vspace{-1em}
\end{minipage}
\hfill
\begin{minipage}[t]{0.30\textwidth}
    \centering
    \vspace{-0.5em}
    \includegraphics[width=0.95\linewidth]{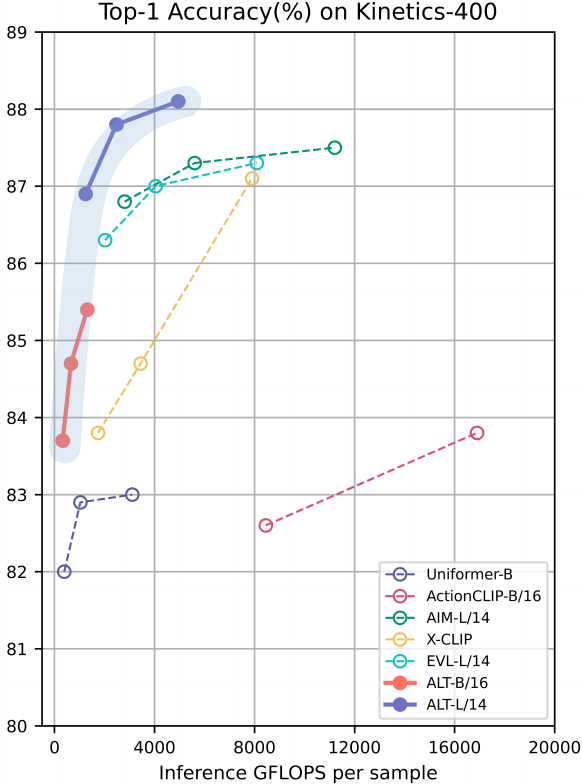}
    \vspace{-1em}
\end{minipage}
\hfill
\quad
\caption{\small 
\textbf{Left:} 
Visualization of visual-semantic correspondences with the tool~\cite{Chen_gscoreCAM22}. For each row: Column (2) visualizes the visual correspondence to text entities generated by the action label; Column (3) visualizes region-aware embeddings under ToMe; Column (4) and (5) show the two of the fine-grained corresponding visual patterns to specific text entities, which are geometrically consistent with Column (3).
\textbf{Right:} 
Visualization of Accuracy v.s. FLOPs performance.
}
\label{fig:align_vis_and_GFLOPs_compare}
\vspace{-1em}
\end{figure*}

\begin{SCtable*}
\centering
\vspace{-1em}
\resizebox{0.795\textwidth}{!}{  
\begin{tabular}{lcccccc|ccccccc}
\toprule
\multirow{2}{*}{Method} & \multirow{2}{*}{Frames} & \multirow{2}{*}{GFLOPs} &\multicolumn{4}{c}{HMDB-51} & \multicolumn{4}{c}{UCF-101}  \\
& & & $K$=2 & $K$=4 & $K$=8 & $K$=16 & $K$=2 & $K$=4 & $K$=8 & $K$=16 \\
\midrule
TSM~\cite{JiTSM}
& 32 & - & 17.5  & 20.9  & 18.4   & 31.0  & 25.3  & 47.0  & 64.4 & 61.0 \\
TimeSformer~\cite{bertasiusSpacetimeAttentionAll2021} 
& 32 & 238 & 19.6  & 40.6  & 49.4   & 55.4  & 48.5  & 75.6  & 83.7 & 89.4 \\
VideoSwin-B~\cite{liuVideoSwinTransformer2022a} 
& 32 & 321 & 20.9  & 41.3  & 47.9   & 56.1  & 53.3  & 74.1  & 85.8 & 88.7 \\
\midrule
ActionCLIP~\cite{wangActionclipNewParadigm2021} 
& 8  & 141 & 55.0  & 56.0  & 58.0   & -     & 80.0  & 85.0  & 89.0 & - \\
X-CLIP-B/16~\cite{niXCLIP} 
& 32 & 658 & 53.0  & 57.3  & 62.8   & 64.0  & 76.4  & 83.4  & 88.3 & 91.4 \\
X-Florence~\cite{niXCLIP}  
& 32 & 2822 & 51.6  & 57.8  & 64.1   & 64.2  & 84.0  & 88.5  & 92.5 & 94.8 \\
ViFi-CLIP~\cite{RasheedViFiCLIP23}
& 32 & 562 & 57.2 & 62.7 & 64.5 & 66.8  & 80.7 & 85.1 & 90.0 & 92.7 \\
ALT-B/16
& 32 & 436 & 64.3 & 66.7 & 70.4 & 74.5 & 93.2 & 95.3 & 96.4 & 97.3    \\
ALT-L/14
& 32 & 1649 & \textbf{68.1} & \textbf{69.7} & \textbf{73.8} & \textbf{79.1} & \textbf{96.0} & \textbf{97.4} & \textbf{98.0} & \textbf{98.1}    \\
\bottomrule
\end{tabular}
\vspace{1em}
\caption{Few-shot comparison: we compare ALTs with previous SOTAs on HMDB-51 and UCF-101. All the models are trained on Kinetics-400, with top-1 accuracies(\%) reported under a single-view inference.}
}
\vspace{1em}
\label{tab:fewshot}
\end{SCtable*}


\begin{SCtable*}
\caption{Zero-shot comparison between ALTs with representative image \& image-language pretraining methods on HMDB-51 \& UCF-101 (left) and Kinetics-600 (right). Pretrained on Kinetics-400, the accuracies(\%) are reported under a single-view inference.} 
\setlength{\tabcolsep}{3.2mm}
\label{tab:zeroshot_all}
\begin{minipage}[t]{0.40\textwidth}
\centering
\resizebox{\textwidth}{!}{  
\begin{tabular}{lccc}
\toprule
Method 
& HMDB-51      & UCF-101  \\
\midrule
ASR~\cite{QianASR} 
& 21.8$\pm$0.9 & 24.4$\pm$1.0 \\
ZSECOC~\cite{JieZSECOC} 
& 22.6$\pm$1.2 & 15.1$\pm$1.7 \\
UR~\cite{YiUR} 
& 24.4$\pm$1.6 & 17.5$\pm$1.6 \\
TS-GCN~\cite{JunyuTSGCN} 
& 23.2$\pm$3.0 & 34.2$\pm$3.1 \\
ER-ZSAR~\cite{ShizheERZSAR} 
& 35.3$\pm$4.6 & 51.8$\pm$2.9 \\
\midrule
ActionCLIP~\cite{wangActionclipNewParadigm2021} 
& 40.8$\pm$5.4 & 58.3$\pm$3.4 \\
X-CLIP-B/16~\cite{niXCLIP} 
& 44.6$\pm$5.2 & 72.0$\pm$2.3 \\
ASU-B/16~\cite{Chen22ASU} 
& 48.1$\pm$2.8 & 75.0$\pm$3.7 \\
ViFi-CLIP~\cite{RasheedViFiCLIP23} 
& 51.3$\pm$0.6 & 76.8$\pm$0.7 \\
ALT-B/16
& 52.9$\pm$1.0 & 79.4$\pm$0.9 \\
ALT-L/14
& \textbf{56.6}$\pm$\textbf{0.8} & \textbf{83.9}$\pm$\textbf{1.1} \\
\bottomrule
\end{tabular}
}
\end{minipage} 
\begin{minipage}[t]{0.39\textwidth}
\centering
\resizebox{\textwidth}{!}{  
\begin{tabular}{lccc}
\toprule
Method 
& Top-1        & Top-5        \\
\midrule
DEVISE~\cite{AndreaDEVISE} 
& 23.8$\pm$0.3 & 51.0$\pm$0.6 \\
ESZSL~\cite{BernardinoESZSL} 
& 22.9$\pm$1.2 & 48.3$\pm$0.8 \\
DEM~\cite{LiDEM} 
& 23.6$\pm$0.7 & 49.5$\pm$0.4 \\
GCN~\cite{PallabiGCN} 
& 22.3$\pm$0.6 & 49.7$\pm$0.6 \\
ER-ZSAR~\cite{ShizheERZSAR} 
& 42.1$\pm$1.4 & 73.1$\pm$0.3 \\
\midrule
ActionCLIP~\cite{wangActionclipNewParadigm2021} 
& 66.7$\pm$1.1 & 91.6$\pm$0.3 \\
X-CLIP-B/16~\cite{niXCLIP} 
& 65.2$\pm$0.4 & 86.1$\pm$0.8 \\
ASU-B/16~\cite{Chen22ASU} 
& 67.6$\pm$0.2 & 87.2$\pm$0.3 \\
ViFi-CLIP~\cite{RasheedViFiCLIP23} 
& 71.2$\pm$1.0 & 92.2$\pm$0.3 \\
ALT-B/16
& 72.7$\pm$0.6 & 91.7$\pm$0.4 \\
ALT-L/14
& \textbf{74.9$\pm$0.4} & \textbf{92.2$\pm$0.3} \\
\bottomrule
\end{tabular}
}
\end{minipage}
\end{SCtable*}

\section{Experiments}
\label{sec:experiments}
\paragraph{Datasets.} Our proposed method is evaluated on four widely used video action recognition datasets: 
Kinetics-400~\cite{kay_kinetics_2017}, 
Kinetics-600~\cite{joaok600},
UCF-101~\cite{KhurramUCF101}, 
HMDB-51~\cite{HildegardHMDB51},
and Something Something V2~\cite{Goyal2017SSv2}(see in supplementary materials).
Kinetics-400 consists of approximately 240k training and 20k validation videos, covering 400 classes, with each clip spanning around 10 seconds. Kinetics-600 is an extension of Kinetics-400, including around 410k training and 29k validation videos for 600 classes. UCF-101 contains 13,320 video clips with 101 classes, and HMDB-51 consists of 7,000 videos with 51 classes. We conduct fully-supervised experiments on Kinetics-400 and Kinetics-600. Additionally, for Kinetics-600, UCF-101, and HMDB-51, we perform few-shot and zero-shot experiments with models pre-trained on Kinetics-400.

\subsection{Fully Supervised Comparison}
\noindent \textbf{Settings.} We conduct fully-supervised experiments on Kinetics-400. Each video clip is sampled with 16 or 32 frames. Two variants of the network, namely ALT-B/16 and ALT-L/14, employ ViT-B/16 and ViT-L/14, respectively. The results on Kinetics-600 and Something-Something-v2~\cite{Goyal2017SSv2} are exhibited in the supplementary materials.

\noindent \textbf{Results.} 
In \cref{tab:k400}, we compare with the state-the-of-art methods on Kinetics-400 with the input resolution 224$\times$224.
Taking 16 sampled frames of each video as input, ALT-B/16 achieves \emph{84.8\%} top-1 accuracy with only 657 GFLOPs. When the input frames increase to 32, ALT-B/16 surpasses the performance of ViViT-H/16~\cite{arnabVivitVideoVision2021a}, which takes more than 30$\times$ computation cost (1308 vs. 48916 GFLOPs). By employing the larger backbone, ALT-L/14 achieves superior performance with \emph{88.1\%} top-1 accuracy among CLIP-400M pretraining works and significant computational advantage ( \emph{0.4\%} higher than AIM~\cite{YangAIM22} but \emph{55\%} fewer GFLOPs). It is noteworthy that the leading method MTV-H~\cite{yanMultiviewTransformersVideo2022} adopts larger-scale pretraining data (70M video-text pairs with about 17B images) and consumes 9$\times$ GFLOPs. 
We attribute the superiority and efficiency of ALT to the seamless coupling of entity-to-region alignments token merging.
As shown in \cref{fig:align_vis_and_GFLOPs_compare} right, we visualize the `\emph{performance v.s. GFLOPs}' relationships of some representative works, where ALT sets new Pareto frontiers.


\subsection{Few-shot Comparisons}

\noindent \textbf{Settings.} 
We evaluate our few-shot experiments on the HMDB-51 and UCF-101 datasets, utilizing ALTs trained on Kinetics-400 data and a text corpus. For the training set construction, we randomly sample 2, 4, 8, and 16 videos from each class, and we set the frame number in each video to either 8 or 32. 
Following the protocols of X-CLIP~\cite{niXCLIP}, we use the first split of the test set for evaluation.

\noindent \textbf{Results.} 
\cref{tab:fewshot} shows the performance comparison on $K$-shot learning. Our method significantly outperforms image-pretrained methods. For instance, when $K$=2, ALT-B/16 surpasses VideoSwin-B~\cite{liuVideoSwinTransformer2022a} by \emph{43.4\%} on HMDB-51 and \emph{39.9\%} on UCF-101. Among the approaches that leverage image-language pretraining: In two-shot scenarios, ALT-B/16 surpasses the previous state of arts by \emph{7.1\%} and \emph{9.2\%} on HMDB-51 and UCF-101, respectively. The lead remains consistent across 16-shot scenarios and continues to expand when switching to ALT-L/14. This showcases the superior generalization capabilities of our paradigm, which establishes reusable entity-to-region alignments.

\subsection{Zero-shot Comparisons}

\noindent \textbf{Settings.} 
For zero-shot evaluation, we train ALTs on Kinetics-400 data and corpus, following the protocol outlined in the \cite{niXCLIP}:  For HMDB-51 and UCF-101, we conducted experiments using the three provided splits. Regarding Kinetics-600, the test set is constructed by randomly selecting 160 categories, three times, from 220 categories that are distinct from those in Kinetics-400. We report results in the format of \textit{``average accuracy $\pm$ standard deviations.''}

\begin{table*}[htb]
\begin{minipage}[t]{0.48\textwidth}
    \vspace{-1em}
    \centering
    \vspace{-0.5em}
    \renewcommand{\arraystretch}{1.25}
    \resizebox{0.98\linewidth}{!}{
    \begin{tabular}{rcccccc}
    \toprule
     & Align. & Corpus. & Region. & Fully. & 2-shot & 0-shot \\
    \midrule
    a. & \twemoji{multiply} & \twemoji{multiply} & \twemoji{multiply} & 81.7 & 53.0 & 72.0\\
    b. & \twemoji{check mark} & \twemoji{multiply} & \twemoji{multiply} & 82.0 & 55.2 & 74.4\\
    c. & \twemoji{check mark} & \twemoji{check mark} & \twemoji{multiply} & 82.2 & 58.1 & 76.9\\
    d. & \twemoji{check mark} & \twemoji{check mark} & \twemoji{check mark} & \textbf{82.8} & \textbf{64.3} & \textbf{79.4}\\
    \bottomrule
    \end{tabular} }
    \vspace{-0.5em}
    \captionof{table}{\small Effect of proposed components. 
    \textbf{\textit{Align}}: `Alignment' before adaptation step; 
    \textbf{\textit{Corpus}}: Utilize action-related text corpus;
    \textbf{\textit{Region}}: Empowered by region-aware visual embeddings. 2-shot and 0-shot are evaluated on HMDB-51 and UCF-101, respectively. 
    }
    \label{tab:components}
\end{minipage}
\hfill
\begin{minipage}[t]{0.48\textwidth}
    \centering  
    \vspace{-1.5em}  
    \renewcommand{\arraystretch}{0.95}
    \resizebox{0.95\linewidth}{!}{
    \begin{tabular}{rcccccc}
    \toprule
     & CA & SA & 1D-Conv & Fully. & 2-shot & 0-shot \\
    \midrule
    a. & \twemoji{multiply} & \twemoji{multiply} & \twemoji{multiply}    & 81.1 & 55.0 & 68.3\\
    b. &\twemoji{check mark} & \twemoji{multiply} & \twemoji{multiply} & 81.8 & 61.3 & 74.6\\
    c. & \twemoji{check mark} & \twemoji{check mark}  & \twemoji{multiply} &82.5 & 62.7 & 76.3\\
    d. & \twemoji{check mark} & \twemoji{multiply} & \twemoji{check mark} & 82.3 & 62.1 & 76.1\\
    e. & \twemoji{check mark} & \twemoji{check mark} & \twemoji{check mark}  & \textbf{82.8} & \textbf{64.3} & \textbf{79.4}\\
    \bottomrule
    \end{tabular} }
    \vspace{-0.5em}
    \captionof{table}{\small Effect of components in the video adapter: 
    \textbf{\textit{CA}}: cross attention module; 
    \textbf{\textit{SA}}: Self attention module;
    \textbf{\textit{1D-Conv}}: 1D convolution module.
    Top-1 Acc. are reported in a single view.
    2-shot and 0-shot are evaluated on HMDB-51 and UCF-101, respectively.} \label{tab:video_adapter_components}
\end{minipage}
\\
\begin{minipage}[t]{0.48\textwidth}
    \vspace{1em}   
    \renewcommand{\arraystretch}{1.2}
    \centering
    \resizebox{0.98\linewidth}{!}{
    \begin{tabular}{rlcccc}
    \toprule
    & Method & \makecell[c]{Top-1 \\Acc.(\%)} & GFLOPs & \makecell[c]{Param.\\(M)} & \makecell[c]{Tunable \\Param.(M)}\\
    \midrule
    a. & Frozen  & 81.4 & 110 & 134 & \textbf{38}\\
    b. & S-adapter~\cite{YangAIM22} & 81.7 & 116 & 138 & 42\\
    c. & SM-adapter~\cite{YangAIM22}  & 81.8  & 123 & 141 & 45\\
    d. & STM-adapter~\cite{YangAIM22}  & 82.5 & 163 & 145 & 49\\
    e. & Finetune  & \textbf{82.8} & \textbf{110} & \textbf{134} & 134\\
    \bottomrule
    \end{tabular} } 
    \vspace{-0.5em} 
    \captionof{table}{\small Effect of different training strategies on the image encoder. \textbf{\textit{S}}: spatial;
    \textbf{\textit{M}}: MLP;
    \textbf{\textit{T}}: temporal; 
    \textbf{\textit{Param.}}: \# parameters. We report Top-1 Acc. on Kinetics-400 under fully-supervised settings.}
    \label{tab:train_strategy}
\end{minipage}   
\hfill
\begin{minipage}[t]{0.48\textwidth}
    \renewcommand{\arraystretch}{1.3}
    \centering
    \vspace{1em}   
    \resizebox{0.96\linewidth}{!}{
    \begin{tabular}{rlcccc}
    \toprule
     & Method & \makecell[c]{Tunable\\Param.(M)} & 2-shot & 8-shot  & 16-shot\\ 
    \midrule
    a. & STM-adapter~\cite{YangAIM22}
    & 49  & 58.1 & 66.9 & 71.0  \\
    b. & V-Finetune & 125  & 64.3 & 70.4  & \textbf{74.5} \\
    \midrule
    c. & VL-prompt\**~\cite{RasheedViFiCLIP23}
    & 0.15(+63) & 63.0 & 69.6 & 72.0   \\ 
    d. & ALT+VL-prompt
    & 0.15+38.4 & \textbf{65.3} & \textbf{71.1} & 74.2   \\ 
    \bottomrule
    \end{tabular}
    }
    \vspace{-0.5em}
    \captionof{table}{\small Analysis on different finetune strategies in HMDB-51 few-shot learning. The base model is ALT-B except \** employs ViFi-CLIP, which additionally pretrains the text encoder(63M paramters).}
    \label{Tab:fewshot_finetune}
\end{minipage}  
\end{table*}

\noindent \textbf{Results.} 
We present the zero-shot results in \cref{tab:zeroshot_all}. ALT-B/16 outperforms ViFi-CLIP by \emph{1.6\%}, \emph{2.6\%}, and \emph{1.5\%} in terms of top-1 accuracy on HMDB-51, UCF-101, and Kinetics-600, respectively. It is noteworthy that our
approach requires \emph{22\%} fewer GFLOPs and ViFi-CLIP unfreezes the CLIP text encoder in the pretraining stage. We attribute the superiority to the utilization of the text corpus, whose factorized and reusable semantics mitigate the difficulty of adapting our model to a new scenario. 


\subsection{Ablation Study}
We employ ALT-B/16 to conduct detailed ablation experiments. By default, the fully-supervised experiments are evaluated on Kinetics-400 with 8 frames per video clip. Taking 32 frames per sample as input, 
The few-shot and zero-shot experiments are conducted on the first split of HMDB-51 and UCF-101, respectively. With 32 frames per video clip, we report results under a single-view inference.

\noindent \textbf{Component analysis.} 
We investigate the effectiveness of the proposed components and report the results in \cref{tab:components}:
(a) We consider X-CLIP-B/16~\cite{niXCLIP} (without text prompts) as the baseline. It incorporates a cross-frame module within the CLIP image encoder and follows the ``align then adapt'' approach. The top-1 accuracy on Kinetics-400 is \emph{81.7\%}.
(b) Then we enhance the baseline with our framework but only using the text embedding of the action name rather than the corpus of entities in Eq.\ref{eq:semantic_attention}. The improvement reveals the effectiveness of the proposed video adapter, which utilizes text embedding to guide the adaption from image embeddings to the final video embedding. 
(c) When we replace the action name with the corpus of entities, the results further increase especially in the few/zero-shot scenarios. The phenomenon indicates using the relevant entities can achieve a better generalization ability
in action recognition. 
(d) We further change the global visual image to region-aware embeddings in the alignment(Eq.\ref{eq:semantic_attention}), fulfilling the `entities-to-regions' alignment. Our approach finally improves the baselines by \emph{0.9\%}, \emph{11.3\%}, and \emph{7.4\%}, respectively.

\noindent \textbf{Video adapter component analysis.}
One important function of our proposed video adapter is to enable modality interactions of established alignments through the cross-attention (CA) modules. To examine its effects, in ~\cref{tab:video_adapter_components}: (a) we initially set up a baseline by substituting the video adapter with four self-attention layers. (b) Then we evaluate the performance with a CA-only video adapter, which improves the baseline significantly, validating the effectiveness of `entity-to-region' alignments. (c)-(e) Additionally, we investigated the effects of 1D-convolution and self-attention (SA) modules which enable spatiotemporal signal communication. The results reveal that both modules are beneficial and compatible with each other.

\noindent \textbf{Training strategy and efficiency of image encoder.}
Our method finetunes the pre-trained image encoder, which influences the final performance as shown in \cref{tab:train_strategy}: 
(a) when freezing the image encoder during training, fewer tunable parameters are required while the accuracy decreases to \emph{81.4\%}.
(b)-(d) Based on the `frozen' setting, we leverage approaches proposed by AIM~\cite{YangAIM22}, which introduces adapters into the image encoder. The performances are improved by stacking spatial, temporal, and MLP adapters in the transformer blocks. 
It is noteworthy that the STM method (d) computes the attention layers twice, therefore significantly increasing the computational complexity of our framework (e) by \emph{48\%}.

\noindent \textbf{Finetune strategy in few-shot scenarios.}
We investigate different finetuning strategies in few-shot learning, and the results are shown in ~\cref{Tab:fewshot_finetune}:
(a) STM-adapter~\cite{YangAIM22} can significantly improve the frozen model baseline (\emph{52.5\%}) with a small number of training parameters,
(b) Finetuning the whole visual branch further improves performance without overfitting.
(c) The recent prompt tuning method,  VL-prompt$^{*}$~\cite{RasheedViFiCLIP23}, demonstrates impressive performance in few-shot learning. 
(d) We adopt the technique in ALT by freezing the backbone and adding ten learnable tokens to each layer of the image \& text encoders. This modification leads to improved accuracies when dealing with lower shots.

\begin{table}[t]
    \centering
    \vspace{-0.5em}
    \centering
    \renewcommand{\arraystretch}{1.2}
    \resizebox{0.98\linewidth}{!}{
    \begin{tabular}{lcccc|c}
    \toprule
    Dataset & Split 1 & Split 2 & Split 3 & Average & Benchmark~\cite{Tong22VideoMAE} \\
    \midrule
    UCF-101 & 95.6 & 95.8 & 96.1 & 95.83 & \textbf{96.1} \\
    HMDB-51 & 73.8 & 73.5 & 74.0 & \textbf{73.77} & 73.3 \\
    \bottomrule
    \end{tabular}}
    \vspace{-0.5em}
    \captionof{table}{\small Linear evaluation on ALT-B/16 (Kinetics-400 pretrained) with Top-1 Acc. reported under a single view. \textit{Benchmark} denotes VideoMAE that fully trained on Kinetics-400 and target datasets.}
    \label{tab:linear_eval}
    \vspace{-1em}
\end{table}

\noindent \textbf{Linear evaluation on learned representations.}
 To assess the quality of the video representations obtained, we conduct linear probe experiments on UCF101 and HMDB-51 datasets with a frozen ALT-B/16. Specifically, the representations of ALT are fixed and fed into a tunable linear classification layer with one-hot label supervision. The top-1 accuracies are presented in \cref{tab:linear_eval} and compared with VideoMAE~\cite{Tong22VideoMAE}, which adopts video reconstruction supervision to learn powerful representations and achieve impressive recognition performance. Notably, our linear-probed results are comparable to a fully-trained VideoMAE, underscoring the discrimination and generalization capabilities of our approach.




\section{Conclusion}
In this paper, we propose a novel method with the ``align before adapt" paradigm for video action recognition.
By leveraging the alignments between local visual appearance and action-related entity semantics, we conduct a better video representation adaption with improved interpretability and generalizability. Our method demonstrates superior performance especially in zero-shot and few-shot scenarios while maintaining low computational costs.


{
    \small
    \bibliographystyle{ieeenat_fullname}
    \bibliography{main}
}

\appendix
\section{Training configuration}
In fully-supervised training, we set the batch size to 256 and adopt the AdamW optimizer with $\beta_1=0.9$ and $\beta_2=0.98$. The learning rate is 8$\times10^{-6}$ for the ViT backbone in the region-aware image encoder and 8$\times10^{-5}$ for the remaining learnable parts. In few-shot experiments, the learning rate of the video adapter is scaled up by ten times, and the batch size is reduced to 64. Regarding the text corpora, we initially constructed a text corpus from ~\cite{kay_kinetics_2017,KhurramUCF101,HildegardHMDB51} with a total number of 913 text entities. 
All the text entities are embedded offline and fixed throughout experiments. It is worthwhile mentioning that when adapting to the motion-heavy dataset Something Something v2~\cite{Goyal2017SSv2}, which predominantly uses action labels in the format of \textit{``action on \textbf{something}"} without specifying instances, we directly collect the action labels as entities.
For data augmentation, we utilize the technique including \emph{RandomFlip, MultiScaleCrop, Mixup, and Label smoothing}, following the manner of X-CLIP~\cite{niXCLIP}.

\section{Text corpus construction}\label{text_corpus_appendix}
In this paper, we propose an automatic pipeline for generating a text corpus.
For each description, (1) we `extract' the relevant action-related units in two approaches: noun and phrase entities extraction with NLTK \& spaCy~\cite{NLTK,spacy2} part of speech (POS) tools;  And ChatGPT~\cite{OpenAIChatGPT22}, where we design a prompt template \textit{``What are identifying visual characteristics, such as object, body parts, scenes, and roles, of a/an \{label\} video action? List them concisely."} (2) We then use the WordNet~\cite{GeorgeWordNet} tool to generate a sequence of explanatory descriptions for each extracted single-word unit. For the extracted phrase unit, we prompt ChatGPT to generate explanatory descriptions with the following templates:
\textit{``Concisely describe what a/an \{phrase unit\} looks like"},
\textit{``Concisely list potential explanations for \{phrase unit\}"},
and \textit{``Concisely explain \{phrase unit\} in one sentence"}.
(3) To determine the most appropriate description for each unit, we employ the Lesk algorithm~\cite{basile2014EnhancedLesk} and T5-based word sense disambiguation~\cite{wahle2021WSDinLM} model according to the action labels. All of these procedures are automated through code.

\definecolor{mistyrose}{rgb}{1.0, 0.89, 0.88}
\definecolor{bubbles}{rgb}{0.89, 0.98, 1.0}
\setcounter{table}{0}
\renewcommand{\thetable}{C\arabic{table}} 
\begin{table}
\centering
\renewcommand{\arraystretch}{1.2}
\resizebox{0.95\linewidth}{!}{
\begin{tabular}{lcccc}
\toprule
Method & Pretrain & Top-1  & GFLOPs & Views     \\
\midrule
MViT-B-24~\cite{fanMultiscaleVisionTransformers2021}
& -           & 83.8    & 1180 & 32$\times$5$\times$1  \\
VideoSwin-L(384$\uparrow$)~\cite{liuVideoSwinTransformer2022a}
& IN-21k     & 85.9    & 25284 & 32$\times$4$\times$3 \\
ViViT-H/16x2 320~\cite{arnabVivitVideoVision2021a}
& JFT-300M   & 83.0    & - & 32$\times$4$\times$3 \\
ViViT-H/16x2~\cite{arnabVivitVideoVision2021a}
& JFT-300M   & 85.8    & 48916 & 32$\times$4$\times$3 \\
TokenLearner-L/10~\cite{MichaelTokenLearner}
& JFT-300M  & 86.3    & 48912 & 32$\times$4$\times$3 \\
Florence(384$\uparrow$)~\cite{LuFlorence}
& FLD-900M  & 87.8    & - & 32$\times$4$\times$3 \\
CoVeR~\cite{BowenCoVeR}
& JFT-3B   & 87.9    & - & 96$\times$1$\times$3 \\
MTV-L~\cite{yanMultiviewTransformersVideo2022}
& JFT-3B   & 85.4  & 18483  & 32$\times$4$\times$3 \\
\rowcolor{bubbles}
MTV-H~\cite{yanMultiviewTransformersVideo2022}
& WTS-17B   & 89.6    & 45537 & 32$\times$4$\times$3 \\
X-CLIP-L/14~\cite{niXCLIP}
& CLIP-400M     & 88.3     & 7896 & 8$\times$4$\times$3  \\
\midrule
ALT-B/16
& CLIP-400M   & 86.1   & 1308 & 32$\times$1$\times$3  \\
\rowcolor{mistyrose}
ALT-L/14
& CLIP-400M    & 88.6     & 4947 & 32$\times$1$\times$3  \\
\bottomrule
\end{tabular} 
}
\vspace{-0.6em}
\caption{\small Fully-supervised comparison on Kinetics-600.}
\label{tab:k600}
\end{table}

\begin{table}[!h]
\centering
\renewcommand{\arraystretch}{1.3}
\resizebox{0.95\linewidth}{!}{
\begin{tabular}{lcccc}
\toprule
Method & Pretrain & Top-1  & GFLOPs & Views     \\
\midrule
MViT-B-24~\cite{fanMultiscaleVisionTransformers2021}
& K-600           & \textbf{69.7}    & 708 & 32$\times$1$\times$3  \\
ViViT-L~\cite{arnabVivitVideoVision2021a}
& IN-21K+K-400   & 65.4    & 11892 & 32$\times$1$\times$3 \\
MTV-B(384$\uparrow$)~\cite{yanMultiviewTransformersVideo2022}
& IN-21K+K-400   & 68.5  & 11160  & 32$\times$3$\times$4 \\
\midrule
EVL-B/16~\cite{linFrozenCLIPModels2022}
& CLIP-400M     & 62.4     & 2047 & 32$\times$1$\times$3  \\
ST-Adapter-B/16~\cite{panSTAdapterParameterEfficientImagetoVideo2022}
& CLIP-400M     & \underline{69.5}     & 1955 & 32$\times$1$\times$3  \\
ALT-B/16
& CLIP-400M    & 68.6     & 1308 & 32$\times$1$\times$3  \\
\bottomrule
\end{tabular} 
}
\vspace{-0.6em}
\caption{\small Fully-supervised comparison on SS-V2.}
\label{tab:ssv2}
\vspace{-1em}
\end{table}

\section{Additional experiments}
\subsection{Fully-supervised experiments on Kinetics-600}
Tab.~\ref{tab:k600} presents the results on Kinetics-600. 
Our ALT-B/16 outperforms MTV-L~\cite{fanMultiscaleVisionTransformers2021} by \emph{0.7\%} by using 32 frames per video with three views. Equipping with a larger backbone, ALT-L/14 achieves \emph{88.6}\% top-1 accuracy with computation consumption of only 4947 GFLOPs, which takes the lead among the methods that adopt similar-level pre-trained models and data.

\subsection{Fully-supervised experiments on Something-Something v2}
The Something-Something V2 (SS-V2) dataset collects more than 220000 video clips that belong to 174 action categories, covering basic human actions with everyday objects. Compared to Kinetics-400, it requires more temporal reasoning. We evaluate our approach on Something-Something V2 under full supervision. 
The accuracies are reported in Tab.~\ref{tab:ssv2}.
Among the CLIP-based works, our method outperforms EVL~\citep{linFrozenCLIPModels2022}, but it is inferior to ST-adapter~\citep{panSTAdapterParameterEfficientImagetoVideo2022}, which utilizes interleaved heavier 3D Convolution modules. We attribute the key to handling such kind of motion-heavy datasets to elaborately designed temporal communication mechanisms, which inspire future directions of our work.

\begin{table}
\centering
\vspace{-1em}
\renewcommand{\arraystretch}{1.2}
\setlength{\tabcolsep}{3.5mm}
\resizebox{0.45\textwidth}{!}{  
\begin{tabular}{lcccccc}
\toprule
Method & $K$=2 & $K$=4 & $K$=8 & $K$=16 \\
\midrule
Vanilla CLIP~\cite{radfordLearningTransferableVisual2021}
& 2.7 & 2.7 & 2.7 & 2.7 \\
ActionCLIP~\cite{wangActionclipNewParadigm2021} 
& 4.1 & 5.8  & 8.4  & 11.1 \\
X-CLIP-B/16~\cite{niXCLIP} 
& 3.9 & 4.5  & 6.8  &10.0 \\
A5~\cite{juPromptingVisualLanguageModels2022}
& 4.4 & 5.1 & 6.1 & 9.7  \\
ViFi-CLIP~\cite{RasheedViFiCLIP23}
 & 6.2 & 7.4 & 8.5 & 12.4 \\
ALT-B/16
& \textbf{6.6} & \textbf{7.7} & \textbf{9.4} & \textbf{12.9}   \\
\bottomrule
\end{tabular}
}
\caption{Few-shot comparison on Something Something V2. All the models are trained on Kinetics-400, with top-1 accuracies(\%) reported under a single-view inference.}
\label{tab:fewshot_ssv2}
\end{table}

\begin{table}[!h]
    \centering
    \centering
    \renewcommand{\arraystretch}{1}
    \setlength{\tabcolsep}{4.5mm}
    \resizebox{0.95\linewidth}{!}{
    \begin{tabular}{lccc}
    \toprule
    Text corpus                       
    & Fully. & 2-shot & 0-shot \\
    \midrule
    $\varnothing$                               
    & 81.7 & 52.8 & 71.6 \\
    ${body}$                           
    & 81.9 & 56.4 & 73.6 \\
    $object$                  
    & 82.5 & 62.3 & 75.8 \\
    ${scene}$           
    & 82.4 & 61.1 & 75.1 \\
    ${motion}$
    & 81.8 & 58.7 & 74.4 \\
    ${All}$
    & \textbf{82.8} & \textbf{64.3} & \textbf{79.4}\\
    \bottomrule
    \end{tabular}
    }
    \vspace{-0.5em}  
    \caption{\small Effect of subcollections of the text corpus.} 
    \label{tab:semantics}
    \vspace{-1em} 
\end{table}

\subsection{Few-shot experiments on Something-Something v2}
In tab.~\ref{tab:fewshot_ssv2}, we compare ALT-B/16 in the SS-V2 few-shot setting along with methods that
adapt the same CLIP-B/16 for videos. We note that ALT-B/16 consistently surpasses these methods across different shot settings, which demonstrates the effectiveness of our video adapter. However, 
We acknowledge that existing approaches, which utilize image-language models and primarily emphasize visual semantics, do not provide satisfactory solutions in \textbf{low-shot} scenarios for datasets that are motion-heavy or instance-agnostic, such as Something Something v2. Instead, we believe that well-designed temporal modules (including other modalities and motion cues extraction) are crucial for achieving higher performance levels, which are deserving of future exploration.

\subsection{Investigation of types text corpus}
To further validate the effect of text corpus, we set a baseline model by replacing aligned embeddings of text entities $\mathbf{Q}_0$ (Eq.~7) with random learnable queries. The result is reported in the first row of Tab.~\ref{tab:semantics} (2-shot and 0-shot experiments take 32 frames per video as input). Moreover, we evaluate the effectiveness of each sub-collection of text corpus by categorizing the text entities into four groups: object, body parts, scenes, and primitive motion. We find that each category is helpful, and the models with all text entities further outperformed the baseline, especially in the 2-shot (+\emph{11.5\%}) and 0-shot (+\emph{7.8\%}) experiments. 
The results reveal that our categorized text entities are complementary to each other, and semantic alignments promise more robust visual representations when facing a severe lack of data.

\begin{table}[!h] 
    \centering
    \renewcommand{\arraystretch}{1}
    \setlength{\tabcolsep}{8mm}
    \resizebox{0.9\linewidth}{!}{
    \begin{tabular}{lcc}
    \toprule
    r & GFLOPs & Fully.(\%)\\
    \midrule
    0    & 141    & 83.1 \\
    4 & 129 & 83.0 \\
    8 & 110 & 82.8 \\
    13 & 86 & 82.2\\
    \bottomrule
    \end{tabular} 
    }
    \caption{Trade-off between efficiency \& accuracy. \textit{r}: the number of tokens to reduce in each transformer block. Results are reported in a single view.} \label{tab:r_reduction}
    \vspace{-1em}  
\end{table}

\subsection{Efficiency and accuracy trade-off}
By default, we set the number of token reductions per block in the image encoder $r$ to 8. 
Here we further investigate the performance of varying $r$. As shown in Tab.~\ref{tab:r_reduction}, our method achieves the highest accuracy in fully-supervised experiments without the token merging strategy (also no region-aware semantic alignment.)
As $r$ increases, the consumption of computing decreases gradually, but so does the accuracy. On balance, $r$=8 is a cost-effective choice, it is worthwhile to explore and validate additional configurations for low-shot scenarios in future studies.

\end{document}

%% file: preamble.tex
\usepackage[dvipsnames]{xcolor}
